\documentclass[acmtog, authorversion]{acmart}
\usepackage{algorithm}
\usepackage{algorithmic}
\usepackage{multirow}
\usepackage{graphicx} % 用于插入图像
\usepackage{subcaption} % 用于创建子图
\usepackage{pifont}

\AtBeginDocument{%
  \providecommand\BibTeX{{%
    \normalfont B\kern-0.5em{\scshape i\kern-0.25em b}\kern-0.8em\TeX}}}

\setcopyright{acmlicensed}
\copyrightyear{2018}
\acmYear{2018}
\acmDOI{XXXXXXX.XXXXXXX}

\begin{document}

\title{BEND: Bagging Deep Learning Training Based on Efficient Neural Network Diffusion}

\author{Jia Wei}
\email{weijia4473@stu.xjtu.edu.cn}
\affiliation{%
  \institution{Xi'an Jiaotong University}
  \streetaddress{Xianning Street}
  \city{Xi'an}
  \state{Shaanxi}
  \country{China}
  \postcode{710127}
}

\author{Xingjun Zhang}
\affiliation{%
  \institution{Xi'an Jiaotong University}
  \streetaddress{Xianning Street}
  \city{Xi'an}
  \country{China}}
\email{larst@affiliation.org}

\author{Witold Pedrycz}
\affiliation{%
  \institution{University of Alberta}
  \city{Edmonton}
  \country{Canada}
}

% \affiliation{%

% \affiliation{%

% \affiliation{%

% \affiliation{%

% \affiliation{%


\begin{abstract}
Bagging has achieved great success in the field of machine learning by integrating multiple base classifiers to build a single strong classifier to reduce model variance. The performance improvement of bagging mainly relies on the number and diversity of base classifiers. However, traditional deep learning model training methods are expensive to train individually and difficult to train multiple models with low similarity in a restricted dataset. Recently, diffusion models, which have been tremendously successful in the fields of imaging and vision, have been found to be effective in generating neural network model weights and biases with diversity. A trained diffusion model can quickly convert input random noise data into valid neural network model weights and biases. We creatively propose a Bagging deep learning training algorithm based on Efficient Neural network Diffusion (BEND). The originality of BEND comes from the first use of a neural network diffusion model to efficiently build base classifiers for bagging. Our approach is simple but effective, first using multiple trained model weights and biases as inputs to train autoencoder and latent diffusion model to realize a diffusion model from noise to valid neural network parameters. Subsequently, we generate several base classifiers using the trained diffusion model. Finally, we integrate these ba se classifiers for various inference tasks using the Bagging method. Resulting experiments on multiple models and datasets show that our proposed BEND algorithm can consistently outperform the mean and median accuracies of both the original trained model and the diffused model. At the same time, new models diffused using the diffusion model have higher diversity and lower cost than multiple models trained using traditional methods. The BEND approach successfully introduces diffusion models into the new deep learning training domain and provides a new paradigm for future deep learning training and inference.
\end{abstract}



% \keywords{Do, Not, Us, This, Code, Put, the, Correct, Terms, for,



\maketitle

\section{Introduction}
Bootstrap Aggregating (Bagging) was proposed by Leo Breiman in 1996 \cite{breiman1996bagging}, which improves the stability and accuracy of a model by training multiple base classifiers separately and aggregating the predictions. It has been effectively solved the high variance problem in machine learning models such as decision trees, random forests \cite{breiman2001random}, support vector machines\cite{kim2002support}, and neural networks \cite{hinton2012improving}, and has a large number of applications in various real-world application areas such as natural language processing \cite{islam2022analyzing}, computer vision \cite{dietterich2000ensemble}, and bioinformatics \cite{diaz2006gene}. Since we entered the era of deep learning, Bagging has been widely used in the form of Dropout \cite{hinton2012improving} at the neural network level to enhance the expressive power of the model. The classical Bagging algorithm needs to integrate multiple base classifiers with diversity, however, today's deep learning models have high training costs \cite{wei2023fastensor}, and the cost of training deep learning-based base classifiers from scratch is very high and unaffordable, how to obtain more diverse learning-based base classifiers at a lower cost is an important challenge.

Currently, diffusion models have been widely used in visual fields such as images and videos, for example, OpenAI's Sora model \cite{cho2024sora} generates videos with realistic texture and movie-quality picture quality, which makes countless researchers feel excited and shocked. The earliest diffusion models can be traced back to nonequilibrium thermodynamics\cite{jarzynski1997equilibrium, sohl2015deep}. The earliest diffusion models were used in the field of image denoising \cite{sohl2015deep}. Subsequently, models such as DDPM \cite{ho2020denoising} and DDIM \cite{song2020denoising} proposed a two-stage diffusion model that includes a forward process from "image to noise" and a reverse process from "noise to image". After continuous optimization of the model structure and training process, the diffusion model has gradually surpassed the mainstream GAN-based model \cite{he2022synthetic,trabucco2023effective} in the field of image generation. Recently, models such as GLIDE \cite{nichol2021glide}, Imagen \cite{saharia2022photorealistic}, and Stable Diffusion \cite{rombach2022high}, as well as the realization of conditional or unconditional diffusion of photo-realistic images, have been widely used in image data enhancement \cite{he2022synthetic, trabucco2023effective} and art creation.

Besides being used in vision domain areas, diffusion models have recently been found to have the potential to generate high-performance model parameters. Wang et al. \cite{wang2024neural}  found commonalities between image diffusion and parameter generation from both data and task perspectives. From a data perspective, both high-quality images and high-performance parameters can be turned into simple distributions, such as Gaussian distributions, by gradually adding noise. From the task perspective, both the inverse process of diffusion modeling and model training can be thought of as transforming from an initial state to a specific distribution. Based on the above similarities and the ability of diffusion models to transform random noise into specific distributions, they proposed p-diff, which uses autoencoder and DDPM to construct a diffusion model that realizes the transformation from a set of noise to a set of model parameters with high performance. Diffusion model-based parameter generation is more efficient and versatile than the traditional model training process, and can quickly generate a large number of model parameters with different expressive capabilities.

The parameter generation method based on diffusion model is a good solution to the demand for the number and diversity of models in the Bagging scenario.  Based on p-diff, we \textbf{innovatively propose the BEND method, which fully utilizes the advantages of the neural network parameter diffusion model to alleviate the training overhead of deep learning in the Bagging scenario and improve the accuracy of the final model}. BEND uses diffusion models to generate the parameters of a given model instead of the traditional process of generating model parameters based on multiple resampling and deep learning training. Subsequently, we propose two simple but effective ways to integrate the inference results of multiple models, sBEND and aBEND. sBEND is a static inference strategy that uses a model integration strategy based on a voting mechanism to fairly count the inference results of each model and select the inference with the most occurrences as the final inference result. In contrast, aBEND is a dynamic inference strategy that also uses a model integration strategy based on a voting mechanism, with the difference that the final chosen strategy result is randomly selected according to the probability of occurrence of each result. sBEND strategy has higher stability, while aBEND strategy has the potential to achieve higher model performance.

We show through experimental results on multiple models and datasets that BEND can consistently meet or even exceed the average and median accuracies of both the generated and original models. The models diffused by BEND have higher diversity compared to the original models, which further improves the performance of Bagging. In addition to this, using the BEND method incurs less training overhead compared to traditional deep neural network training methods, as long as the number of base classifiers trained is more than 3. Finally, we also analyze the ablation of hyperparameters such as the number of original models, the number of integrated models, and the number of enhanced layers.
% ACM's consolidated article template, introduced in 2017, provides a
% consistent \LaTeX\ style for use across ACM publications, and
% incorporates accessibility and metadata-extraction functionality
% necessary for future Digital Library endeavors. Numerous ACM and
% SIG-specific \LaTeX\ templates have been examined, and their unique
% features incorporated into this single new template.

% If you are new to publishing with ACM, this document is a valuable
% guide to the process of preparing your work for publication. If you
% have published with ACM before, this document provides insight and
% instruction into more recent changes to the article template.

% The ``\verb|acmart|'' document class can be used to prepare articles
% for any ACM publication --- conference or journal, and for any stage
% of publication, from review to final ``camera-ready'' copy, to the
% author's own version, with {\itshape very} few changes to the source.
\section{Background}
\subsection{Diffusion Model}
Diffusion models are sequential hidden variable models inspired by thermodynamic diffusion \cite{sohl2015deep,rombach2022high,wang2024neural}. The training process of the diffusion model can be categorized into two stages: forward noise addition and reverse denoising. Forward noise addition gradually adds Gaussian noise to the original image, and the reverse denoising stage starts with random noise to train a Gaussian transformation through a Markov chain.

\subsubsection{Forward noise addition}

given a sample (real image) $x_0 \sim q(x)$, Gaussian noise is gradually added over $T$ time steps to get different stages of noise addition samples $x_1,x_2,...x_t,...,x_T$. The iterative formula for this stage is shown below,

\begin{align}
    q(x_t|x_{t-1}) &= \mathcal{N}(x_t;\sqrt{1-\beta_t}x_{t-1},\beta_t\mathbf{I}) \\
    q(x_{1:T}|x_0) &= \prod_{t=1}^{T} q(x_t|x_{t-1})
\end{align}
where $q$ represents the forward noise addition process and $\mathcal{N}$ represents the Gaussian noise parameterized by $\beta_t$, and $\mathbf{I}$ represents the identity matrix.

\subsubsection{Reverse denoising}
The reverse denoising process is the inverse of the forward denoising process, which removes noise from $x_t$ by training a denoising neural network step by step through T steps to get a clear image.The iterative formula for the reverse denoising process is shown below,
\begin{align}
    p_\theta(x_{t-1}|x_t) &= \mathcal{N}(x_{t-1};\mu_\theta(x_t,t),\beta_t\mathbf{I}) \\
    p_\theta(x_{0:T}) &= \prod_{t=1}^{T} p_\theta(x_{t-1}|x_t)
\end{align}
where $p$ represnets the reverse denoising process, $\beta_t\mathbf{I}$ is the fixed covariance of the Gaussian transitions, and the learnd mean $\mu_\theta(x_t,t)$ of the Gaussian transitions is defiend below,
\begin{equation}
    \mu_\theta(x_t,t) = \frac{1}{\sqrt{\alpha_t}}(x_t-\frac{\beta_t}{\sqrt{1-\tilde{\alpha}_t}}\epsilon_\theta(x_t,t))
\end{equation}
Ho et al. \cite{} define $\epsilon_\theta(\cdot)$ is a neural network trained to process a noisy sample $x_t$ and predict added noise, $\alpha_t = 1 - \beta_t$ and  $\tilde{\alpha}_t = \prod_{s=1}^{t} \alpha_t$.The denoising neural network in the reverse denoising process is optimized by the standard negative log-likelihood:
\begin{equation}
    \mathcal{L}_{dm} = \mathcal{D}_{KL}(q(x_{t-1}|x_t,x_0)||p_\theta(x_{t-1}|x_t))
\end{equation}
where ${D}_{KL}(\cdot||\cdot)$ denoted the Kullback-Leribler (KL) divergence, which is an asymmetric measure of the difference between two probability distributions.

\subsection{Deep learning training}
Deep learning training is the process of iteratively optimizing deep neural network models by methods such as mini-batch gradient descent. The outstanding performance of deep learning models in computer vision, natural language processing, etc. mainly comes from iteratively updating the parameters of neural network models with hundreds or even thousands of epochs using a large amount of data. With the continuous updating of the network structure (from fully connected neural networks to convolutional neural networks to transformer neural networks) and the deepening of the network depth and width, training a network model to reach convergence often requires days or even weeks of training using high-end GPU clusters.

\begin{figure}
    \centering
    \includegraphics[width=0.5\textwidth]{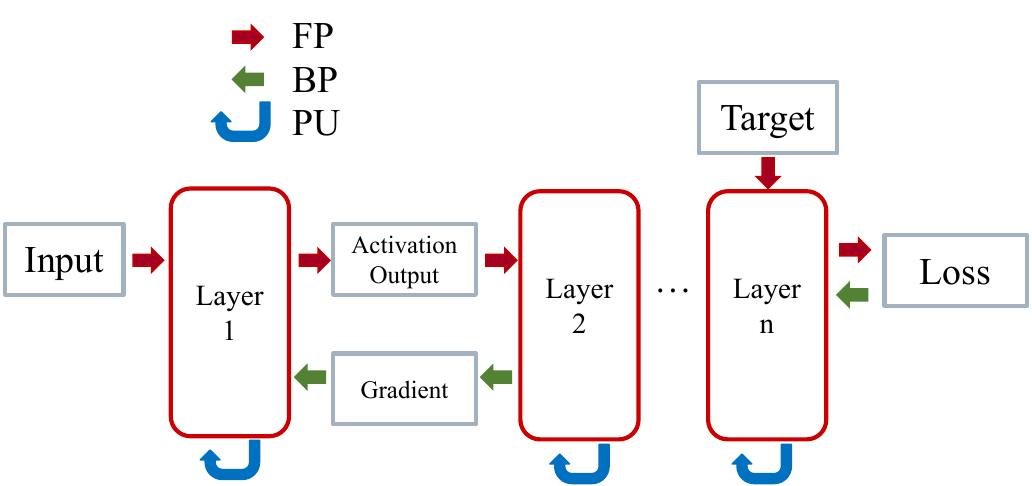}
    \caption{Deep Learning Training Process}
    \label{DNN}
\end{figure}

Specifically, as shown in Figure \ref{DNN}, deep learning training can be divided into three stages: forward propagation (FP), backward propagation (BP), and parameter update (PU). In the FP phase, starting from the preprocessed input data, the weights and biases of the current layer are utilized to perform the given layer operation computation to obtain the activation output of this layer. Then the activation output of the current layer is used as the input of the next layer, and the weights and deviations of the next layer are used to complete the operations of the next layer until the last layer yields the loss value of this iteration based on the given loss function and target result. (2) In the BP phase, starting from the last layer, the loss gradient of the current layer input and the current layer weights and deviations is iteratively computed up to the input layer. (3) Finally, in the PU stage, all activation outputs and their gradients are cleared and all model weights and biases are updated using optimizers such as SGD \cite{Yan2022}, ADAM \cite{Kingma2015}, ADADELTA \cite{zeiler2012adadelta}. The process is repeated until the model reaches a given number of training sessions or other convergence conditions. The computation of each layer of a typical feedforward neural network during forward propagation is shown in Eq. \ref{update}. $\mathcal{X}^{l}$ represents the input (i.e., the input images or feature maps, the input dimension is $K \times N$) of the $l+1$ layer, $\mathcal{X}^{l+1}$ represents the output (the output dimension is $K \times Q$) o of the $l+1$ layer, $\mathcal{W}^{l+1}$ and $\mathbf{b}^{l+1}$  (the weights and biases dimension are $N \times Q$ and $K$) represent the weights and bias of the $l+1$ layer, $\sigma$ represents the activation function, and $l$ represents the number of layers.
\begin{equation}
    \label{update}
    \mathcal{X}^{(l+1)} = \sigma(\mathcal{X}^{(l)}) \mathcal{W}^{(l+1)} + \mathbf{b}^{(l+1)}
\end{equation}

Different types of neural networks (e.g., convolutional neural networks, recurrent neural networks, etc.) and different layers may have different layer-by-layer formulas, but the basic principles are similar. 

\section{BEND}
\begin{figure*}
    \centering
    \includegraphics[width=\textwidth]{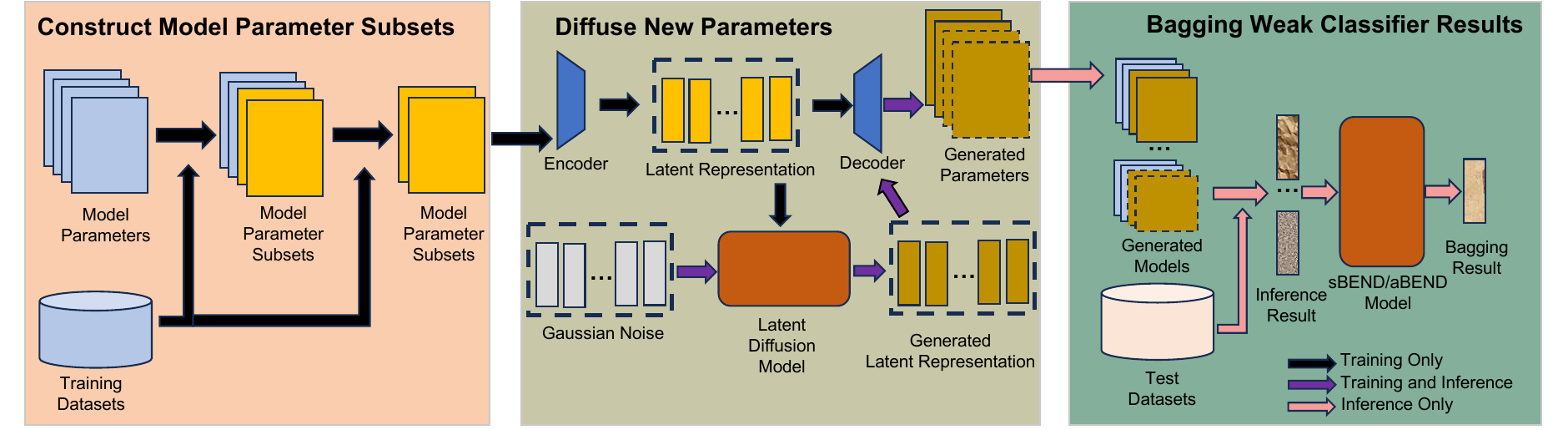}
    \caption{BEND Framework. Our approach consists of three processes, named  Construct Model Parameter Subsets, Diffuse New Parameters,  and  Bagging Weak Classifier Results. Construct Model Parameter Subsets only appears during training. It delivers a subset of trained model parameters. Diffuse New Parameter emerges in both the training and inference phases. Diffuse New Parameter emerges in both the training and inference phases. In the inference phase, it uses noise to diffuse the new model parameters. Bagging Weak Classifier Results occurs only in the inference phase. It ensembles multiple predictions using the sBEND or aBEND methods.}
    \label{framework}
\end{figure*}

We propose a Bagging Deep Learning Training Framework (BEND) based on efficient neural network diffusion.The overall structure of BEND is shown in Fig. \ref{framework}. BEND consists of three main parts.
(1) Construct Model Parameter Subsets. We first select a subset of model parameters (several model layer weights and biases) that need to be diffused, and then train the model using the given dataset to obtain a set of high-performance parameter subsets. (2) Diffusion New Parameters. We first train neural network parameter diffusion using a p-diff-like architecture. The diffusion model consists of two parts: parameter autoencoder and generation. We first use the subset of model parameters obtained in the first stage to train an autoencoder to realize the interconversion from model parameters to latent representation. Subsequently, we train a standard hidden diffusion model to realize the conversion from random noise to a specific latent representation. Finally, we use the trained diffusion model to diffuse from random noise to obtain a new subset of model parameters. (3) Bagging Weak Classifier Results. We construct multiple base classifiers by combining the obtained subset of model parameters with other layers of the frozen model. The results obtained from the base classifiers are then aggregated using the two aggregation strategies provided by sBEND or aBEND within BEND.
\subsection{Construct Model Parameter Subsets}
The process of constructing a subset of model parameters is essentially a combination of training the model from scratch and fine-tuning it. First the entire model needs to be trained from scratch for a number of epochs to reach basic model convergence. Then, the parameters of the non-model parameter subset are frozen, and then the model subset is fine-tuned for $k$ epochs, after each epoch, the model parameter subset $s\theta_k$ is saved once, and a family of sets $FS={ s\theta_1 , s\theta_2 , ...,s\theta_{k-1}, s\theta_k}$ is constructed for subsequent diffusion model training. It is worth noting that the training process of the model from scratch is optional, and it is also possible to start with the pre-trained model parameters, simply fine-tune a few epochs before freezing the non-model parameter subset parameters, and then use the same approach to obtain $FS$.
\subsection{Diffusion New Parameters}
The diffusion model diffuses new parameters. We first train the diffusion model using a subset of the model parameters in $FS$. The diffusion model consists of the autoencoder model, which is responsible for the conversion of model parameters to latent representations, and the DDPM standard latent diffusion model, which is responsible for the conversion from noise to latent representations. We first train the autoencoder using $FS$. in this process, each subset of model parameters in $FS$ is flattened to a one-dimensional vector, and the whole $FS$ is converted to a $d \times k$ matrix $M$, where d is the dimension of the aggregated subset of model parameters. As shown in Fig. X, assuming that our model parameter subset contains a convolutional layer of dimension $[512 \times 512 \times 3 \times 3]$ (with corresponding model parameter weights and biases $w_c$ and $b_c$) and a BN layer of dimension [512] (with corresponding model parameter weights and biases $w_b$ and $b_b$), the final result of its transformation is a vector of length 4719616. When the number of model parameters used to train the autoencoder itself, k, is 200, $M$. In addition to this, to enhance the robustness and expressiveness of the model, we also added random noise to the input data during training. The autoencoder we use contains a 4-layer encoder and decoder, and uses the minimum mean square error (MSE) regarded as the loss function.

\begin{equation}
    \mathcal{L}_{\text{MSE}} = \frac{1}{k}\sum_{1}^{k}||s\theta_k-s\theta'_k||^2
    \label{mse}
\end{equation}
where $s\theta'_k$ is the k-th model reconstructed subset of parameters.

Subsequently, since the size of the model parameter subset matrix $M$ may become very large and difficult to compute when multiple model layers and large models are selected, we use the obtained latent representation rather than the model parameter subset itself to train the DDPM diffusion model.The updating process of the DDPM model is described as follows,
\begin{equation}
    \theta \leftarrow \theta - \nabla_\theta||\epsilon - \epsilon_\theta(\sqrt{\overline{\alpha}_t}z_k^0+\sqrt{1-\overline{\alpha}_t}\epsilon,t)||^2
    \label{ddpm}
\end{equation}
where $t$ represents the time step, which is usually denoted by time embedding, $\overline{\alpha}_t$ is a hyperparameter representing the noise strength at each step, $\epsilon$ represents the added Gaussian noise, and $z_k^0$ indicates the input latent representation data.
\subsection{Bagging Weak Classifier Results}
\label{bagging}
First, we predicted the test data using each weak classifier separately. Subsequently, we aggregate the prediction results obtained from each weak classifier using sBEND or aBEND, and finally obtain the final prediction results.

The sBEND computation procedure is shown in Algorithm \ref{sBEND}, which is consistent with the classical Bagging algorithm. Firstly, the test results of each weak classifier for each test sample are counted. Then, the category with the most prediction results is selected as the final prediction result.
\begin{algorithm}
\caption{sBEND}
\label{sBEND}
\begin{algorithmic}[1]
\REQUIRE Weak Classifier Results $R$, $R$ is a $k*n$ matrix. Target $T$, $T$ is a $n$-dimensional vector.
\ENSURE Inference Results $I$ and Accuracy $a$.
\FOR{ $i$ in range($R$.shape[1])}
    \STATE $column$ = $R[:,i]$
    \STATE $values$, $counts$ = Unique($column$) \COMMENT{Count the number of times each value in the column appears}
    \STATE $most\_frequent\_value$ = $values[\text{Argmax}[$counts$]]$
    \STATE $I[i]$ =  $most\_frequent\_value$
\ENDFOR
\STATE $matching\_elements\_count$ = Sum($I$ == $T$)
\STATE $a$ = $matching\_elements\_count$ / $n$
\RETURN $I$, $a$
\end{algorithmic}
\end{algorithm}

The aBEND computational procedure is shown in Algorithm \ref{aBEND}. Similar to the aBEND algorithm, sBEND also first counts the test results of each weak classifier for each test sample. Then, aBEND selects one of the results as the final prediction based on the probability of occurrence of each prediction.
\begin{algorithm}
\caption{aBEND}
\label{aBEND}
\begin{algorithmic}[1]
\REQUIRE Weak Classifier Results $R$, $R$ is a $k*n$ matrix. Target $T$, $T$ is a $n$-dimensional vector.
\ENSURE Inference Results $I$ and Accuracy $a$.
\FOR{ $i$ in range($R$.shape[1])}
    \STATE $column$ = $R[:,i]$
    \STATE $values$, $counts$ = Unique($column$) \COMMENT{Count the number of times each value in the column appears}
    \STATE $most\_frequent\_value$ = $values[\text{Argmax}[$counts$]]$ 
    \IF{len($values$) > 1}
        \STATE $probability$ = $counts$ / Sum($counts$)
        \STATE $most\_frequent\_value$ = Choice($values$, $probability$) \\ \COMMENT{Select value according to the probability of occurrence of the values}
    \ENDIF
    \STATE $I[i]$ =  $most\_frequent\_value$
\ENDFOR
\STATE $matching\_elements\_count$ = Sum($I$ == $T$)
\STATE $a$ = $matching\_elements\_count$ / $n$
\RETURN $I$, $a$
% \STATE $S \gets \text{getsizeof}(T)$ \COMMENT{Tensor Size Calculation}
% \IF{Customized Dictionary or Customized Format is not None \COMMENT{Dictionary or Tool Determination}} 
%     \STATE $F \gets$ Customized Dictionary Specific Format or $F \gets$ Customized Format
%     \RETURN $F$
% \ENDIF
% \IF{$FlushDic$ is True}
% 	\STATE Flush all Dics
% \ENDIF
% \IF{User use  “w+r” mode and $S$ in $WRDic$ \COMMENT{Write-Read Mode Check}}
% 	\STATE $F \gets$ $WRDic$ Specific Format with $S$ as key
% 	\RETURN $F$
% \ENDIF
% \IF{User use “r” mode and $S$ in $RDic$ and len($RDic$[$S$] ) == 5 \COMMENT{Read Mode Check}}
%     \STATE $F \gets$ $RDic$ Specific Format with with $S$ as key
%     \RETURN $F$
% \ENDIF
% \IF{$S$ in $WDic$ \COMMENT{Write Mode Process and Write Dic Update}}
% 	\IF{len($WDic$[$S$] ) == 5}
% 		\STATE $F \gets$ Get value from $WDic$ with $S$ as key
% 	\ELSIF{len($WDic$[$S$] ) == 1}
% 		\STATE $F \gets$ “npy” and update $WDic$
% 	\ELSIF{len($WDic$[$S$] ) == 2}
% 		\STATE $F \gets$ “pt” and update $WDic$
% 	\ELSE
% 		\STATE $F \gets$ “DALI” and update $WDic$ then Choose best method and update $WDic$ again
% 	\ENDIF
%         \RETURN $F$
% \ENDIF
% \STATE $F \gets$ “cpy” and update $WDic$
% \RETURN $F$
\end{algorithmic}
\end{algorithm}

\begin{figure*}[ht]
    \centering
    \begin{subfigure}[b]{0.3\textwidth}
        \centering
        \includegraphics[width=\textwidth]{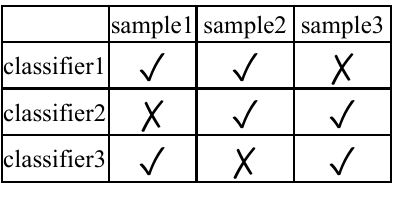}
        \caption{IoU=0}
        \label{i1}
    \end{subfigure}
    \hfill % 添加空白来分隔子图
    \begin{subfigure}[b]{0.3\textwidth}
        \centering
        \includegraphics[width=\textwidth]{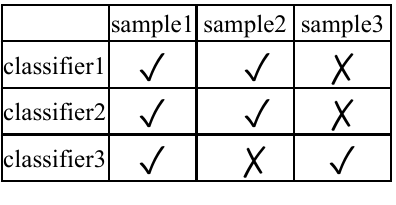}
        \caption{IoU=0.667}
        \label{i2}
    \end{subfigure}
    \hfill % 添加空白来分隔子图
    \begin{subfigure}[b]{0.3\textwidth}
        \centering
        \includegraphics[width=\textwidth]{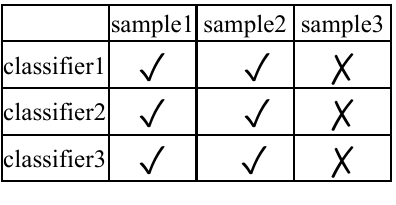}
        \caption{IoU=1}
        \label{i3}
    \end{subfigure}
    \caption{An example of Bagging.There are three samples (sample1, sample2 and sample3) and three classifiers (classifier1, classifier2 and classifier3). \ding{51} means correctly classified and \ding{55} indicates incorrectly classified.}
    \label{example}
\end{figure*}

A simply explained example of the difference between sBEND and aBEND is shown in Figure \ref{example}. Suppose we have three different base classifiers (classifier1, classifier2 and classifier3) that have the same classification accuracy of 66.7\%. We express the dissimilarity of the three models using the Intersection over Union (IoU) of the incorrect predictions of the three models defined by Equation x. The smaller the IoU, the greater the dissimilarity. We discuss the performance of sBEND and aBEND for IoUs of 0, 0.667, and 1, respectively.

As shown in Table \ref{iou}, the performance of sBEND and aBEND is highly correlated with the dissimilarity between the base classifiers. When the dissimilarity between base classifiers is very high and the truth is always in the hands of the majority, sBEND achieves the optimal classification accuracy. For example, in the scenario with IoU = 0 (Figure \ref{i1}), sBEND improves the accuracy of the weak classifier from 66.7\% to 100\%. When the weak classifier dissimilarity is very low, the accuracy improvement from both sBEND and aBEND decreases. For example, in the scenario with IoU=1 (Figure \ref{i3}), neither sBEND nor aBEND can bring performance improvement. When there is some dissimilarity among the base classifiers and there is a situation where the truth is in the hands of a few, there is a probability that aBEND will get an accuracy that exceeds that of sBEND. For example, in the scenario with IoU = 0.667 (Figure \ref{i2}), aBEND has a 22\% probability of outperforming sBEND and has the same mathematical expectation as sBEND.

\begin{table}[]
\caption{Performance of sBEND and aBEND under different IoUs}
\label{iou}
\begin{tabular}{cccc}
\hline
\textbf{Probability} & \textbf{Method} & \textbf{Accuracy} & \textbf{IoU}           \\ \hline
1                    & No-bagging      & 66.7\%            & \multirow{6}{*}0                      \\
1                    & \textbf{sBEND}           & \textbf{100\%}                  &                        \\
0.3                  & aBEND           & 100\%                  &                        \\
0.44                 & aBEND           & 66.7\%              &                        \\
0.22        & aBEND           & 33.3\%              &                        \\
0.04                 & aBEND           & 0\%                  &                        \\ \hline
1                    & No-bagging      & 66.7\%              & \multirow{5}{*}{0.667} \\
1                    & sBEND           & 66.7\%                  &                        \\
0.22                 & \textbf{aBEND}           & \textbf{100\%}                  &                        \\
0.56                 & aBEND           & 66.7\%             &                        \\
0.22                 & aBEND           & 33.3\%             &                        \\ \hline
1                    & No-bagging      & 66.7\%             & \multirow{3}{*}{1}     \\
1                    & sBEND           & 66.7\%             &                        \\
1                    & aBEND           & 66.7\%            &                        \\ \hline
\end{tabular}
\end{table}

\section{Experiments}
We provide a comprehensive evaluation of the BEND algorithm and ablation analysis of key hyperparameters in this section. We first provide a detailed description of the hardware and software environment used for the experiments, the model and dataset, and the selection of hyperparameters in Section \ref{set}. Then, Section \ref{acc} evaluates the test set accuracy of BEND. Section \ref{eff} compares the efficiency of BEND with traditional training methods. Section \ref{diver} analyzes the diversity of model parameters generated by the diffusion model and the original model parameters obtained by training.Section \ref{abl} performs ablation analysis for the number of subsets of model parameters ($k$), the number of base classifiers obtained by diffusion ($m$), and the type of layers of training, respectively.

\subsection{Experimental Setting}
\label{set}
The main hardware and software environments used in this paper are shown in Table \ref{hardware}.
\begin{table}
  \caption{Hardware and Software Environment in the Experiments}
  \label{hardware}
  % \small
  \begin{tabular}{cc}
    \hline
    \textbf{Hardware}  & \textbf{Version}\\
    \hline
    CPU & Intel(R) Xeon(R) Silver 4210R\\
    GPU & NVIDIA A100\\
    NVMe SSD & Samsung MZ-V8V1T0BW \\
    \hline
    \textbf{Software}  & \textbf{Version}\\
    CUDA & 12.2 \\
    Pytorch & 2.0.1 \\
    Numpy &  1.24.4\\
    \hline
  \end{tabular}
\end{table}

We conducted experimental validation on two datasets, Cifar-10 and Cifar-100, using ResNet18, ResNet50, RegNetX400M and RegNetY400M, respectively. Unless otherwise stated, the number of subsets of model parameters $k$ used in our experiments is 200, and the number of base classifiers obtained by diffusion $m$ is 100. the number of epochs trained from scratch when training with the Cifar-10 and Cifar-100 datasets are 100 and 200, respectively.

\subsection{Test Set accuracy}
\label{acc}
As shown in Table \ref{accuracy}, we trained and reasoned on the Cifar-10 and Cifar-100 datasets using ResNet-18, ResNet-50, and RegNet models, respectively. We compared the results of our proposed sBEND, aBEND with the other 7 baselines. All our experiments were run for 10 times and the mean values were recorded. In addition to this, for aBEND we marked thestandard deviation values of the ten runs (aBEND was executed 10 times in each round of experiments, i.e., aBEND was run 50 times in total). By comparing the results of the experiments we can draw the following conclusions:
\begin{enumerate}
    \item sBEND and aBEND achieve comparable or even better test set accuracies to the original model mean and median. This implies that the diffusion model learns an efficient distribution of model parameters and is able to efficiently diffuse model parameters with high-performance performance, starting from random noise.
    \item  sBEND is more stable than aBEND, and the prediction results of each round are not affected by random numbers. At the same time, aBEND has the potential to further improve the test set accuracy. aBEND tends to obtain a larger maximum test set accuracy than sBEND.
    \item sBEND and aBEND consistently obtained mean and median values compared to the generated model. This result demonstrates the effectiveness and general applicability of the bagging method when used for deep learning inference.
    \item sBEND and aBEND outperform the mean and median accuracy of both the original and generated models across multiple models and datasets. This result illustrates that the BEND method is able to effectively utilize the parameters of the generated models to jointly improve the accuracy of the final test set with good generalization.
\end{enumerate}

\begin{table*}[]
\renewcommand{\arraystretch}{1.25} %设置行间距为1.5倍
\caption{Test Set Accuracy of the BEND Method vs. the Baseline Method. We choose the maximum (\textbf{max}), mean (\textbf{mean}), and median (\textbf{med}) values of the generated model as well as the maximum (\textbf{orig\_max}), mean (\textbf{orig\_mean}), and median (\textbf{orig\_med}) values of the original model and the potential maximum (\textbf{potential}) values of the Bagging method as baselines to compare with our proposed sBEND (\textbf{sBEND}) and aBEND (\textbf{aBEND}). \textbf{potential} means that a sample is considered to be predicted correctly as long as any one of the base classifiers predicts it correctly, and this value represents a theoretical upper bound on the accuracy of this set of base classifiers.}
\label{accuracy}
\begin{tabular}{cccccccccc}
\hline
{\color[HTML]{000000} \textbf{sBEND}} & {\color[HTML]{000000} \textbf{aBEND}} & {\color[HTML]{000000} \textbf{max}} & {\color[HTML]{000000} \textbf{mean}} & {\color[HTML]{000000} \textbf{med}} & {\color[HTML]{000000} \textbf{orig\_max}} & {\color[HTML]{000000} \textbf{orig\_mean}} & {\color[HTML]{000000} \textbf{orig\_median}} & {\color[HTML]{000000} \textbf{potential}} & {\color[HTML]{000000} \textbf{task}}  \\ \hline
{\color[HTML]{000000} 94.58}          & {\color[HTML]{000000} $\mathbf{94.59\pm 0.02}$}& {\color[HTML]{000000} 94.64}        & {\color[HTML]{000000} 94.58}         & {\color[HTML]{000000} 94.58}        & {\color[HTML]{000000} 94.69}              & {\color[HTML]{000000} 94.58}              & {\color[HTML]{000000} 94.59}                 & {\color[HTML]{000000} 94.71}              & {\color[HTML]{000000} res18-cifar10}  \\
{\color[HTML]{000000} 76.51}          & {\color[HTML]{000000} $\mathbf{76.52 \pm 0.03}$} & {\color[HTML]{000000} 76.62}        & {\color[HTML]{000000} 76.16}         & {\color[HTML]{000000} 76.52}        & {\color[HTML]{000000} 76.79}              & {\color[HTML]{000000} 76.49}              & {\color[HTML]{000000} 76.49}                 & {\color[HTML]{000000} 79.11}              & {\color[HTML]{000000} res18-cifar100} \\
{\color[HTML]{000000} 94.32}          & {\color[HTML]{000000} $\mathbf{94.33 \pm 0.03}$}          & {\color[HTML]{000000} 94.37}        & {\color[HTML]{000000} 94.32}         & {\color[HTML]{000000} 94.32}        & {\color[HTML]{000000} 94.44}              & {\color[HTML]{000000} 94.28}              & {\color[HTML]{000000} 94.28}        & {\color[HTML]{000000} 94.87}              & {\color[HTML]{000000} res50-cifar10}  \\
{\color[HTML]{000000} \textbf{77.49}} & {\color[HTML]{000000} $77.45 \pm 0.05$}          & {\color[HTML]{000000} 77.61}        & {\color[HTML]{000000} 77.31}         & {\color[HTML]{000000} 77.49}        & {\color[HTML]{000000} 77.72}              & {\color[HTML]{000000} 77.43}              & {\color[HTML]{000000} 77.42}                 & {\color[HTML]{000000} 84.1}               & {\color[HTML]{000000} res50-cifar100} \\
{\color[HTML]{000000} 94.75} & {\color[HTML]{000000} $94.75 \pm 0.03$}          & {\color[HTML]{000000} 94.78}        & {\color[HTML]{000000} 94.73}         & {\color[HTML]{000000} 94.75}        & {\color[HTML]{000000} 94.93}              & {\color[HTML]{000000} \textbf{94.77}}     & {\color[HTML]{000000} 94.77}                 & {\color[HTML]{000000} 95.61}              & {\color[HTML]{000000} regx-cifar10}   \\
{\color[HTML]{000000} 75.91}          & {\color[HTML]{000000} $75.82 \pm 0.05$}          & {\color[HTML]{000000} 75.96}        & {\color[HTML]{000000} 75.77}         & {\color[HTML]{000000} 75.91}        & {\color[HTML]{000000} 76.32}              & {\color[HTML]{000000} 76.03}              & {\color[HTML]{000000} \textbf{76.04}}        & {\color[HTML]{000000} 80.48}              & {\color[HTML]{000000} regx-cifar100}  \\
{\color[HTML]{000000} 94.42}          & {\color[HTML]{000000} $94.44 \pm 0.02$}          & {\color[HTML]{000000} 94.49}        & {\color[HTML]{000000} 94.42}         & {\color[HTML]{000000} 94.42}        & {\color[HTML]{000000} 94.6}               & {\color[HTML]{000000} \textbf{94.48}}     & {\color[HTML]{000000} 94.48}                 & {\color[HTML]{000000} 95.14}              & {\color[HTML]{000000} regy-cifar10}   \\
{\color[HTML]{000000} \textbf{76.97}}          & {\color[HTML]{000000} $76.95 \pm 0.03$}          & {\color[HTML]{000000} 77.05}        & {\color[HTML]{000000} 76.94}         & {\color[HTML]{000000} 76.95}        & {\color[HTML]{000000} 77.17}               & {\color[HTML]{000000} 76.82}     & {\color[HTML]{000000} 76.82}                 & {\color[HTML]{000000} 78.66}              & {\color[HTML]{000000} regy-cifar100}  \\ \hline
\end{tabular}
\end{table*}
\renewcommand{\arraystretch}{1} %重置行间距为默认值

\subsection{Training Efficiency}
\label{eff}
The total time $t_{diff}$ for the diffusion model to produce a weak classifier consists of the following components:
\begin{enumerate}
    \item Training the original subset of model parameters ($t_{org}$).
    \item Training the autoencoder ($t_{ate}$).
    \item Training the DDPM ($t_{ddpm}$).
    \item Generating a new subset of model parameters ($t_{gen}$).
\end{enumerate}
The formula for $t_{diff}$ is as following:
\begin{align}
    t_{diff} &= t_{org} + t_{ate} + t_{ddpm} + t_{gen} \\
             &= (k_{pre} + k) \times t_{orge} + t_{ate} + t_{ddpm} + m \times t_{sgen}
\end{align}
where $k_{pre}$ represents the number of epochs required to train a model to convergence, $t_{orge}$ is the time required to train the model for one epoch, $t_{ate}$ and $t_{ddpm}$ represent the time required to train the autoencoder and ddpm respectively, and $t_{sgen}$ represents the time required to use the diffusion model to diffuse a weak classifier using the diffusion model.

Traditional deep learning training overhead is the product of the number of base classifiers and the number of epochs trained for each weak classifier and the computation time for each epoch. The formula for $t_{trad}$ is as following:
\begin{equation}
    t_{trad} = m \times k_{pre} \times t_{orge}
\end{equation}

Taking the example of training the cifar-10 dataset using the ResNet50 model, the relevant parameter values we used and calculated in our experiments are shown in Table \ref{TE}.
\begin{table}
  \caption{Values of Key Parameters of Training Efficiency}
  \label{TE}
  % \small
  \begin{tabular}{cccccc}
    \hline
    $\mathbf{k_{pre}}$  & $\mathbf{k}$(s) & $\mathbf{t_{orge}}$(s) & $\mathbf{t_{ate}}$(s) & $\mathbf{t_{ddpm}}$(s) & $\mathbf{t_{sgen}}$(s)\\
    \hline
    100 & 200 & 28 & 931.39 & 857.14 & 2.35 \\
    \hline
  \end{tabular}
\end{table}
Therefore, we can calculate the conditions under which the diffusion efficiency is higher than the traditional method as:
\begin{equation}
    m > 3.64
\end{equation}
This result implies that as many as more than 3 base classifiers are generated, the BEND method obtains a higher training efficiency than the traditional method.

\subsection{Generated Model Parameter Diversity}
\label{diver}
We use the differences in the predictions of each classifier with respect to other classifiers to indicate the diversity among different classifiers. The diversity $d_i$ within the generated model and the original model is calculated as follows:
\begin{equation}
    d_i = \frac{\sum_{i=1}^{m} \sum_{j=1}^{m} \sum_{k=1}^{n} 
    \begin{cases} 
        1 & \text{if } p_i[k] \neq p_j[k] \ \text{and} \ i \neq j, \\
        0 & \text{else.}
\end{cases}}{m-1}
\end{equation}
The diversity $d_o$ between the generated model and the original model is calculated as follows:
\begin{equation}
    d_o = \frac{\sum_{i=1}^{m} \sum_{k=1}^{n} 
    \begin{cases} 
        1 & \text{if } p_i[k] \neq q_i[k], \\
        0 & \text{else.}
\end{cases}}{m}
\end{equation}
where $m$ represents the number of base classifiers, $n$ stands for the number of data samples, and $p$ and $q$ denote the classification results of the generated and original base classifiers.

\begin{figure}
    \centering
    \includegraphics[trim=120 430 0 150, width=0.5\textwidth]{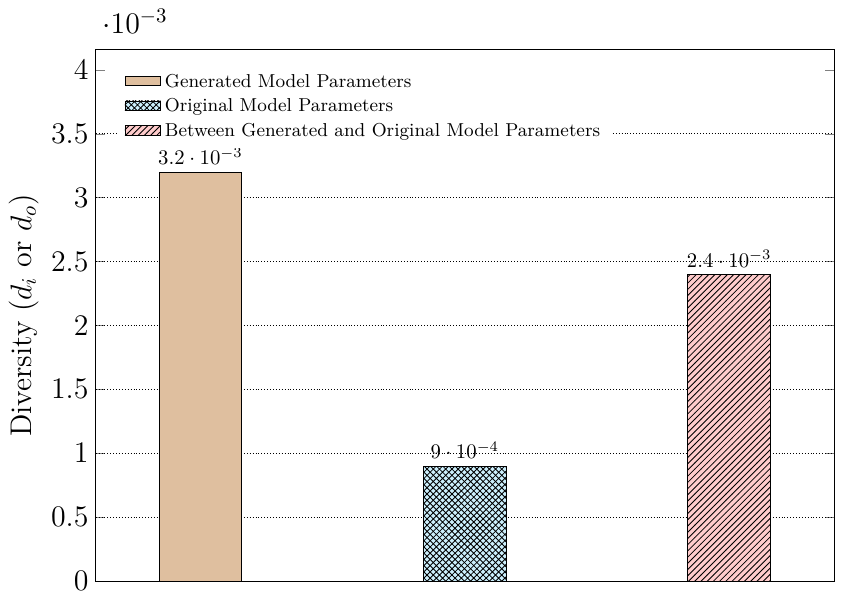}
    \caption{Diversity of Different Model Parameters}
    \label{div}
\end{figure}

As shown in Fig. \ref{div}, we have computed the diversity among the generated model parameters, the diversity among the original model parameters, and the diversity between the generated model parameters and the original model parameters, respectively. The experimental results show that the model parameters obtained from the diffusion model have a higher diversity between them, even higher than between the model obtained from diffusion and the original model. As we discuss in Section \ref{bagging}, the higher diversity of the diffusion models means that they are more suitable for Bagging.

\begin{figure*}
    \centering
    \begin{subfigure}[b]{0.49\textwidth}
        \centering
        \includegraphics[trim=120 400 0 100, width=\textwidth]{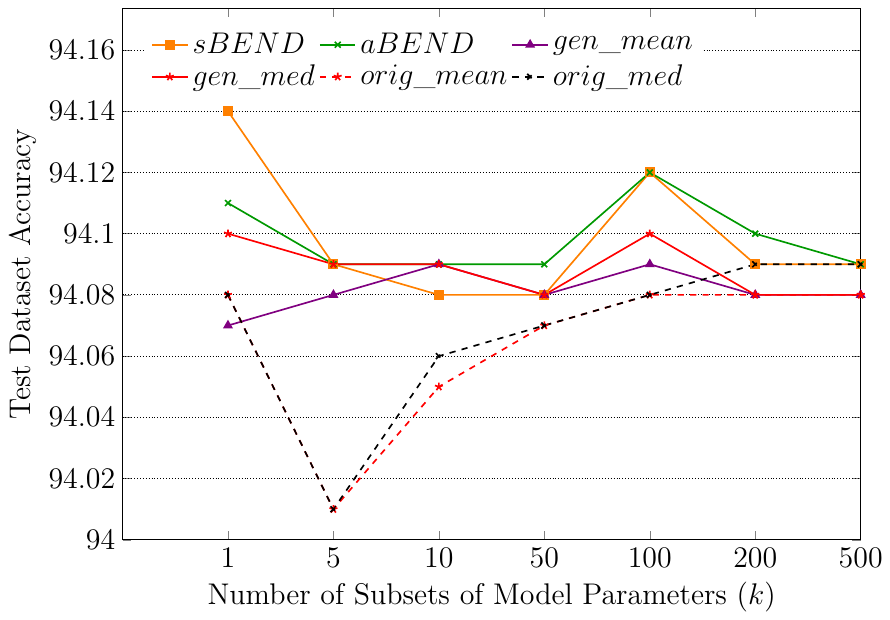}
        \caption{Test Set Accuracy vs. the Number of Subsets of Model Parameters ($k$)}
        \label{k}
    \end{subfigure}
    \hfill % 添加空白来分隔子图
    \begin{subfigure}[b]{0.49\textwidth}
        \centering
        \includegraphics[trim=120 400 0 100, width=\textwidth]{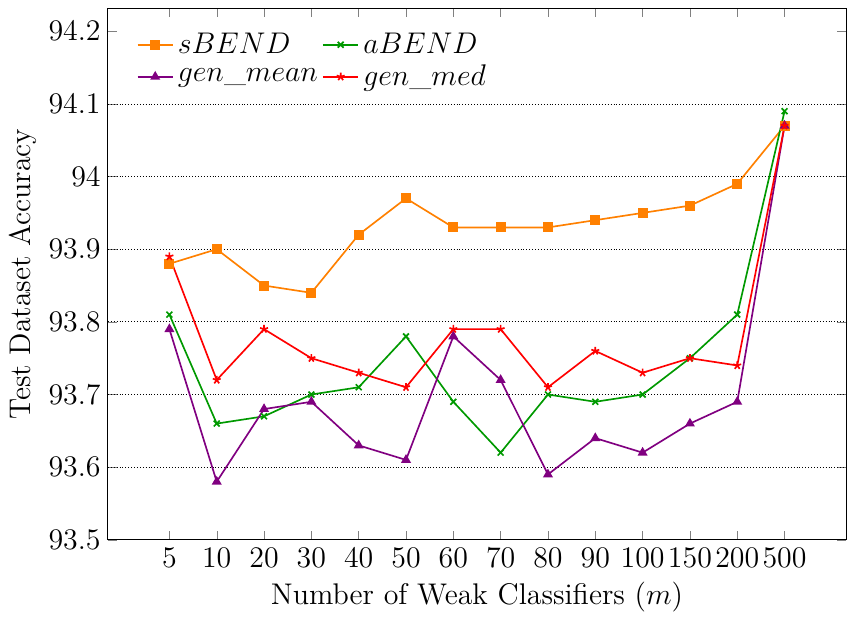}
        \caption{Test Set Accuracy vs. the Number of Diffused base classifiers($m$)}
        \label{m}
    \end{subfigure}
    \caption{Test Set Accuracy vs. the Number of Subsets of Model Parameters ($k$) and the Number of Diffused base classifiers ($m$)}
    \label{kandm}
\end{figure*}
\subsection{Ablation Analysis}
\label{abl}
In this section, ablation analyses are performed for the number of subsets of model parameters ($k$), the number of base classifiers obtained by diffusion ($m$), and the type of training layers, respectively, in order to obtain the effect of different hyperparameters on the BEND test set accuracy.

\subsubsection{The Number of Subsets of Model Parameters ($k$)}
In this section, the test set accuracy of the ResNet-50 model derived for the Cifar-10 dataset using 100 generated base classifiers (m=100) with seven different numbers of subsets of model parameters chosen from k=1 to k=500.

As shown in Figure \ref{k}, the accuracy of the model test set obtained by the BEND method is consistently higher than the mean ( $gen\_{mean}$) and median ( $gen\_{med}$) of the generated model and the mean ( $orig\_{mean}$) and median ( $orig \_{med}$) of the original model, regardless of the variation of $k$.

At the same time, we have observed that the size of k has little effect on model accuracy, which means that we only need to train to produce a small subset of model parameters to train a well-performing diffusion model.

\subsubsection{The Number of Diffused base classifiers ($m$)}
In this section, the test set accuracy of the ResNet-50 model derived for the Cifar-10 dataset with the same 200 subsets of model parameters (k=200) by choosing different numbers of base classifiers from m=5 to m=500.

As shown in Figure \ref{m}, model accuracy essentially increases with the number m of base classifiers generated. sbend performs more steadily than abend and consistently outperforms the mean and median.

\subsubsection{Types and Positions}
In order to further explore the influence of the layers and types where the subset of model parameters are located on the models obtained from the final diffusion as well as on the Bagging results, we conducted experimental analyses for three typical different types of layers in ResNet-50. We chose two convolutional layers, four BN layer combinations, and one fully connected layer to analyze, respectively. The subset of model parameters for each convolutional layer contains the weights of one convolutional layer, and there are no Bias for the parameters of the convolutional layers of the ResNet-50 model.The subset of model parameters for each BN layer contains the weights and biases of two BN layers.The subset of model parameters for the FC layer contains the weights and biases of the FC layer.

The experimental results are shown in Table \ref{layer}, and we find that most of the model parameters of the types and positions achieve comparable or even better performance than the original model parameters. For different types of model parameter subsets have different effects on the final Bagging performance, in general, the BN layer has better accuracy and stability than the convolutional (Conv) and fully-connected ( FC) layers. The main reason is that the hidden variable model is sensitive to the input size of the diffusion model, and when the parameter size is too large (e.g., layer3.5.conv3 has more than 200,000 parameters), the model obtained by diffusion is less effective. In addition to this, for the same type of layers, deeper diffusion layers give better accuracy than shallower diffusion layers. This observation is consistent with p-diff's findings \cite{wang2024neural} that for BN layers, the parameters of the shallower layers are more likely to accumulate errors during training than those of the deeper layers.

\begin{table}[]
\caption{Test Set Accuracy Comparison of the BEND Method with the Original Model for Different Layer Selections.}
\label{layer}
\renewcommand{\arraystretch}{1.25} %设置行间距为1.5倍
\begin{tabular}{ccccc}
\hline
{\color[HTML]{000000} \textbf{Layer}}                              & sbend                & abend                & orig          & param \\ \hline
{\color[HTML]{000000} layer4.2.bn3,layer4.2.bn2   }          & 94.32                & $94.33 \pm 0.02$                & 94.28                & 5120         \\
{\color[HTML]{000000} layer4.0.bn3,layer4.0.bn2   }          & 94.27                & $94.23 \pm 0.02$                 & 94.14                & 5120         \\
{\color[HTML]{000000} layer3.5.bn3,layer3.5.bn2   }          & 94.25                & $94.12 \pm 0.12$                & 94.25                & 2560         \\
{\color[HTML]{000000} layer3.0.bn3,layer3.0.bn2   } & 94.10                & $94.13 \pm 0.05$                & 94.14                 & 2560         \\
{\color[HTML]{000000} layer1.0.conv1.weight}                    & 93.88                & $93.09 \pm 0.14$                & 93.91                & 4096         \\
{\color[HTML]{000000} layer3.5.conv3.weight}                       & 84.01 & $83.72 \pm 0.11$   & 94.01 & 262144       \\
{\color[HTML]{000000} fc}                                          & 93.94                & $89.17 \pm 0.23$                 & 94.27                & 20490        \\ \hline
\end{tabular}
\end{table}
\renewcommand{\arraystretch}{1} %设置行间距为1.5倍
\section{Related Work}
\subsection{Bagging}
Bagging as a common integrated learning method has been widely used in various machine learning algorithms and many fields. Leo et al. \cite{breiman1996bagging} firstly proposed the Bagging algorithm and introduced its basic principle, implementation method, and the application effect on different models. Subsequently, they combined the Bagging algorithm with a decision tree algorithm to propose a Random Forest model \cite{breiman2001random}, which increases the diversity among the base decision trees by randomly selecting a subset of features at each split. Hinton et al. \cite{hinton2012improving} link Bagging and Dropout methods, arguing that Dropout can be viewed as an extreme form of Bagging, which is equivalent to sampling a large number of different "sub-networks" from the original network, training them, and then approximating the average of the predicted results during inference. Dietterich et al. \cite{dietterich2000ensemble} used the Bagging method to integrate multiple Convolutional Neural Networks to improve the performance in the tasks of Image Classification and Target Detection. Diaz-Uriarte et al. \cite{diaz2006gene} used Bagging to integrate several different machine learning models to improve the accuracy of disease diagnosis. Smaida et al. \cite{smaida2020bagging} integrated three different convolutional neural network models for ocular disease monitoring. Wu et al. \cite{ wu2017data}, on the other hand, combined the k-means method and bagging neural networks together to improve the accuracy of short-term wind power prediction. All of these methods have expensive computational overheads as all machine learning models must be trained extensively before integrating multiple machine learning models using bagging.

\subsection{Synthetic Data}
Synthesized data is currently being used to optimize various task-heavy areas in computer vision \cite{antoniou2017data,tran2017bayesian,zheng2017unlabeled}. In early research, synthetic data based on GANs were heavily used for representation learning \cite{jahanian2021generative}, inverting images \cite{zhang2020image}, semantic segmentation \cite{zhang2021datasetgan} and training classifiers \cite{wickramaratne2021conditional, haque2021ec}. Recently, diffusion models \cite{li2023your,hertz2022prompt,peebles2023scalable,ho2022imagen} have made remarkable achievements in the field of image generation and have replaced GANs as a hotspot for generative data research. These methods are based on the thermodynamic diffusion approach which has a similar pathway as GANs \cite{zhu2017unpaired,brock2018large}, VAEs \cite{kingma2013auto,razavi2019generating} and flow-based models \cite{dinh2014nice,rezende2015variational}. Saharia and Rombach et al. proposed Imagen \cite{saharia2022photorealistic} and Stable Diffusion \cite{rombach2022high} models to improve the quality of images synthesized by diffusion models, respectively. Song et al. \cite{song2020denoising} and Bao et al. \cite{bao2022analytic} worked on improving the sampling speed of the models. He et al. \cite{he2022synthetic} and Trabucco et al. \cite{trabucco2023effective} used the image data generated by the diffusion model for a few-shot image classification task. And Wang et al. \cite{wang2024neural} proposed for the first time to use the diffusion model to synthesize model parameters instead of images, expanding the application scope of the diffusion model. To the best of our knowledge, this paper is the first to discuss how to utilize models obtained by diffusion as well as to analyze the performance of integrating these models.
\section{Conclusion and Discussion}
In this paper, we propose BEND, a new paradigm for deep learning training for Bagging using diffusion neural network models. BEND is the first to use diffusion models for Bagging. BEND is able to efficiently train and diffuse high-quality neural network model parameters and effectively integrate them for inference tasks. We propose two variants, sBEND and aBEND, to meet the different needs of different users for higher stability and accuracy. We demonstrate through extensive experiments that the models diffused by BEND have similar or even better accuracy and more diversity than the models trained by traditional methods. At the same time, the experimental results also show the effectiveness, robustness, and generalizability of bagging composed of diffused base classifiers. BEND provides a new mindset and novel paradigm for the cooperation between diffusion models and deep learning models.

BEND has realized the combination of Bagging based on diffusion models and supervised training for deep learning. In the future, we will further explore the integration of learning methods, diffusion modeling, and further deep integration of self-supervised and unsupervised training. In addition, we plan to apply BEND to multimodal domains such as natural language processing, speech recognition, and recommender systems. Besides, diffusion modeling for deep neural network models is still in its infancy, and how to improve the availability, effectiveness, and cost-efficiency of diffusion modeling for network model parameters remains a key challenge.

\bibliographystyle{ACM-Reference-Format}
\bibliography{panda}

%%
% %% If your work has an appendix, this is the place to put it.
% \appendix









\end{document}